\documentclass[11pt]{article}
\usepackage[margin=1in]{geometry}

\usepackage{amsmath,amssymb,amsthm}
\usepackage{booktabs}
\usepackage{multirow}
\usepackage{xcolor}
\usepackage[expansion=false]{microtype}
\usepackage{graphicx}
\usepackage{enumitem}
\usepackage{algorithm}
\usepackage{algpseudocode}
\usepackage[numbers,sort&compress]{natbib}
\usepackage[colorlinks=true,linkcolor=blue!50!black,citecolor=blue!50!black,urlcolor=blue!50!black]{hyperref}
\hypersetup{pdftitle={CAPSkip: Cancellation-Aware Prefix Skipping for Selective Backpropagation},pdfauthor={Heejin Choi},pdfsubject={Preprint}}

\newcommand{\R}{\mathbb{R}}
\newcommand{\pp}{\,pp}
\newcommand{\ci}[2]{[#1,\,#2]}
\newtheorem{proposition}{Proposition}

\theoremstyle{definition}
\newtheorem{definition}{Definition}
\newtheorem{example}{Example}

\title{CAPSkip: Cancellation-Aware Prefix Skipping\\ for Selective Backpropagation}
\author{Heejin Choi\\
  Yonsei University \quad Sooil Development \quad NCreate\\
  \texttt{heejin.share@gmail.com}}
\date{}

\begin{document}
\maketitle

\begin{abstract}\sloppy
Per-step selective back-propagation forwards every example of a mini-batch but back-propagates only some of them.
Most rules rank examples by loss or gradient magnitude, but such scalar scores ignore cancellation between per-example gradients: the gradient removed by a set can be much smaller than its individual magnitudes suggest.
We introduce CAPSkip (Cancellation-Aware Prefix Skipping). Among the examples the model already classifies correctly, CAPSkip skips the longest prefix, in increasing order of last-layer gradient norm, whose \emph{vector sum} stays within a relative budget $b$ of the batch's last-layer gradient.
In exact arithmetic this bounds the change of the raw final-layer gradient, using rank-one contributions already available from the forward pass; the bound does not extend to deeper layers, optimizer updates, or the training trajectory.
Because the cumulative distortion is not monotone in the prefix length, CAPSkip scans all prefixes; a factorised Gram matrix makes the scan cheap and keeps it on the device.
With activation reuse and selected-row backward, CAPSkip fine-tunes DeiT-S on CIFAR-100 for 103 epochs in $0.576\times$ the time of 103-epoch dense training, and is $+0.25$\pp{} \ci{+0.08}{+0.42} above an approximately time-matched 59-epoch dense run ($n=3$).
\end{abstract}

\section{Introduction}

A common way to cut the cost of training is to spend less computation on examples that contribute little to the update.
\emph{Per-step} methods decide inside each mini-batch: they forward all $B$ rows, score them, and back-propagate only a subset~\citep{jiang2019selectivebackprop,katharopoulos2018notall,loshchilov2015online}.
\emph{Per-epoch} methods such as InfoBatch~\citep{qin2024infobatch} and KAKURENBO~\citep{nguyen2023kakurenbo} decide before the forward, from scores stored in earlier epochs, and save both passes.
Common rules in both families score examples individually by loss or gradient magnitude.
The quantity that matters for the update, however, is the \emph{sum} of the skipped gradients.
Per-example gradients of correctly classified examples can be small without being aligned, and their sum can be much smaller than the sum of their norms, so a magnitude-based score misjudges how much of the update is actually removed.

This paper studies a selection rule that measures the skipped gradient directly, in the one place where it is exactly and cheaply available: the final affine layer.
For cross-entropy, the gradient of example $i$'s loss with respect to the final layer's weights and bias is the outer product of the logit residual $h_i$ and the augmented penultimate feature $[\phi_i,1]$.
Both factors are by-products of the forward pass.
We sort the currently-correct examples by the norm of this outer product and skip the \emph{largest} number $k^*$ of them whose summed contribution has relative norm at most $b$.
Because summed vectors cancel, the relative distortion of a prefix is not monotone in its length; stopping at the first prefix that violates the budget under-selects, and magnitude-sum budgets are only conservative sufficient conditions.

\paragraph{Contributions.}
\begin{enumerate}[leftmargin=*,itemsep=1pt,topsep=2pt]
\item An ideal cancellation-aware selection rule (CAPSkip) with an exact-arithmetic per-step bound on the raw last-layer gradient, a full-prefix scan that returns the maximal feasible prefix of the chosen order, and a worked example of why first-violation stopping and magnitude budgets select less (\S\ref{sec:method}).
\item A device-only implementation: all prefix norms come from a 2-D cumulative sum of a factorised Gram matrix $(h_i\!\cdot\! h_j)(\phi_i\!\cdot\!\phi_j+1)$, without materialising per-example gradients for the Gram calculation or synchronising the selector with the host (\S\ref{sec:impl}).
\item A weighted form for composition with outer re-weighting pruners, with a counter-example showing that the unweighted rule can violate the budget on the weighted raw gradient (\S\ref{sec:weighted}).
\item An empirical study with pre-registered decision rules and deliberately narrow claims: an approximately time-matched accuracy gain in one fine-tuning task, an unresolved difference in from-scratch training, a mixed result for composition with InfoBatch, a learning-rate confound we measure explicitly, and the measured cost of executing the rule with and without activation reuse (\S\ref{sec:exp}).
\end{enumerate}

\section{Method}\label{sec:method}

\subsection{Setting and last-layer contributions}
Consider a classifier with logits $z_i = W\phi_i + c \in \R^C$, where $\phi_i\in\R^F$ is the penultimate feature of example $i$, and a mini-batch of $B$ rows trained with the mean loss $L=\frac{1}{B}\sum_{i=1}^B \ell_i$, $\ell_i = \mathrm{CE}(z_i, t_i)$ with a (possibly label-smoothed) target $t_i$.
Write $\tilde\phi_i=[\phi_i;1]$ and $\tilde W=[W\;c]$.
The exact per-example gradient of $\ell_i/B$ with respect to $\tilde W$ is
\begin{equation}
P_i \;=\; h_i\,\tilde\phi_i^{\top},\qquad h_i = \tfrac{1}{B}\bigl(\mathrm{softmax}(z_i)-t_i\bigr),
\qquad \|P_i\|_F = \|h_i\|\,\|\tilde\phi_i\|.
\label{eq:P}
\end{equation}
The full last-layer gradient is $V_0=\sum_{i=1}^B P_i$.
If a set $D$ of rows is removed from the backward, the selected objective $\frac1B\sum_{i\notin D}\ell_i$ has last-layer gradient exactly $V_0-\sum_{i\in D}P_i$ (the normalisation stays $1/B$).
The decomposition requires only that $h_i$ and $\phi_i$ be computed by the forward pass; the models used here (GroupNorm ResNet, LayerNorm ViT) have no cross-example coupling, so per-example gradients are also well defined for deeper layers.

\subsection{The selection rule}
\begin{definition}[CAPSkip]
Let $\mathcal{C}=\{i: \arg\max_c z_{ic} = y_i\}$ be the rows the model currently classifies correctly, $m=|\mathcal{C}|$.
Order them by $\|P_i\|$ ascending, $\pi_1,\dots,\pi_m$, and let $S_k=\sum_{j=1}^k P_{\pi_j}$ ($S_0=0$).
For $\|V_0\|_F>0$, define $D_k=\|S_k\|_F/\|V_0\|_F$. If $V_0=0$, define $D_k=0$ when $S_k=0$ and $D_k=+\infty$ otherwise, so only zero-sum prefixes are feasible.
For a budget $b\ge 0$, CAPSkip drops the rows $\pi_1,\dots,\pi_{k^*}$ from the backward, where
\begin{equation}
k^* \;=\; \max\{\,k\in\{0,1,\dots,m\} : D_k \le b\,\}.
\label{eq:kstar}
\end{equation}
\end{definition}
The maximum in \eqref{eq:kstar} is taken over \emph{all} prefixes (a full scan), not up to the first $k$ with $D_k>b$.
The main comparisons use $b=0.09$; it was selected on validation data in the from-scratch CIFAR-100 setting and transferred unchanged to fine-tuning.
Algorithm~\ref{alg:lacr} gives the practical epsilon-relaxed implementation, whose distinction from this ideal definition is stated below.

\begin{algorithm}[t]
\caption{Practical CAPSkip step (batch of $B$ rows, budget $b$, $\epsilon=10^{-12}$)}\label{alg:lacr}
\begin{algorithmic}[1]
\State Forward the batch once; keep logits $z$, penultimate features $\phi$, and the activations needed for backward.
\State $h \gets (\mathrm{softmax}(z)-t)/B$;\quad $\mathcal{C}\gets\{i:\arg\max z_i = y_i\}$;\quad $r_i\gets \|h_i\|\sqrt{\|\phi_i\|^2+1}$
\State $\pi \gets \operatorname{argsort}(r_i \text{ if } i\in\mathcal{C} \text{ else } +\infty)$ \Comment{candidates first, ascending}
\State $G \gets (h_\pi h_\pi^\top)\odot(\phi_\pi\phi_\pi^\top + 1)$ \Comment{$G_{ab}=P_{\pi_a}\!\cdot P_{\pi_b}$}
\State $s \gets [0,\ \operatorname{diag}(\operatorname{cumsum}_{\text{cols}}(\operatorname{cumsum}_{\text{rows}}(G)))]$ \Comment{$s_k=\|S_k\|^2$, $k=0..B$}
\State $D_k \gets \sqrt{\max(s_k,0)}/(\|V_0\|+\epsilon)$;\quad $k^*\gets\max_k\, k\cdot[k\le|\mathcal{C}|]\cdot[D_k\le b]$ \Comment{device-only}
\State Back-propagate $\frac1B\sum_{i\notin\{\pi_1..\pi_{k^*}\}}\ell_i$ through the kept rows only; optimizer step.
\end{algorithmic}
\end{algorithm}

\subsection{Properties}
The propositions below concern the ideal rule in Definition~1, exact arithmetic, and the raw final-affine-layer gradient before clipping or optimization.

\begin{proposition}[Existence and exact last-layer bound]\label{prop:bound}
$k^*$ always exists, since $D_0=0\le b$.
Removing the prefix changes the raw final-layer gradient by $\Delta=-S_{k^*}$, with $\|\Delta\|_F\le b\|V_0\|_F$. For $V_0\ne0$, this gives $\|\Delta\|_F/\|V_0\|_F=D_{k^*}\le b$; if also $b<1$, the angle between the dense and selected gradients is at most $\arcsin b$. For $V_0=0$, the selected gradient remains zero and its angle is undefined.
\end{proposition}
\begin{proof}
Immediate from $V_0-(V_0-S_{k^*})=S_{k^*}$ and \eqref{eq:kstar}. For the angle, the set $\{V_0+\Delta:\|\Delta\|\le b\|V_0\|\}$ is a ball of radius $b\|V_0\|$ around $V_0$, seen from the origin under half-angle $\arcsin b$.
\end{proof}
This is an exact-arithmetic statement about the ideal rule: $P_i$ is the true per-example last-layer gradient.
It says nothing about the gradients of deeper layers, which can change by more or less than $b$, and nothing about the optimisation trajectory over many steps.

\begin{proposition}[Full scan versus first violation]\label{prop:scan}
Let $k_{\mathrm{fv}}=\min\{k\in\{0,\dots,m-1\}:D_{k+1}>b\}$, with $k_{\mathrm{fv}}=m$ if no prefix violates the budget.
Then $k_{\mathrm{fv}}\le k^*$, and the inequality can be strict.
The feasible set $\{k:D_k\le b\}$ is in general not downward closed: a feasible prefix may contain an infeasible shorter one.
\end{proposition}
\begin{proof}
$k_{\mathrm{fv}}$ is feasible, so $k_{\mathrm{fv}}\le k^*$. Example~\ref{ex:nonmono} gives $k_{\mathrm{fv}}=0<k^*=2$ with $k=1$ infeasible.
\end{proof}

\begin{proposition}[Magnitude budgets are sufficient, not necessary]\label{prop:mag}
$\|S_k\|\le\sum_{j\le k}\|P_{\pi_j}\|$. Hence every $k$ with $\sum_{j\le k}\|P_{\pi_j}\|\le b\|V_0\|$ is feasible, and the largest such $k_{\mathrm{mag}}$ satisfies $k_{\mathrm{mag}}\le k^*$.
\end{proposition}
\begin{proof}
Triangle inequality.
\end{proof}
Two caveats keep these statements honest.
First, $k^*$ is maximal among prefixes \emph{of the norm-ascending order}; it is not the largest subset with $\|\sum_{i\in D}P_i\|\le b\|V_0\|$, which is a combinatorial problem we do not solve.
Second, because the feasible set is not downward closed, any post-processing that shortens the dropped prefix (for instance, returning rows to the backward so that the number of kept rows is a multiple of a kernel tile) can break the budget unless the shorter prefix is itself re-checked; choosing among feasible prefixes only is safe.

\begin{example}[Non-monotone distortion; constructed, not data]\label{ex:nonmono}
Take three candidates whose contributions, already in norm-ascending order, are $(0.9,0)$, $(-1,0)$ and $(0,2)$, and let the other rows contribute $(0.1,8)$, so $V_0=(0,10)$.
With $b=0.05$ the absolute budget is $0.5$.
The prefix norms are $\|S_0\|=0$, $\|S_1\|=0.9$ (violates), $\|S_2\|=0.1$ (feasible again), $\|S_3\|\approx 2.0025$ (violates).
First-violation stopping keeps every candidate ($k=0$); the full scan drops two ($k^*=2$).
A magnitude budget would need $0.9+1=1.9\le 0.5$ for $k=2$ and also selects $k=0$.
\end{example}

\subsection{Weighted composition with an outer pruner}\label{sec:weighted}
An outer per-epoch pruner such as InfoBatch~\citep{qin2024infobatch} skips some rows before the forward and up-weights others: rows whose stored loss is below the mean are skipped with probability $r$ (we use $r=0.5$), and the surviving ones are weighted by $1/(1-r)$.
With the outer weights detached from differentiation, the forwarded-batch objective is $\frac1B\sum_i w_i\ell_i$, whose raw last-layer gradient is $\sum_i w_iP_i$.
The inner rule must therefore be applied to the weighted contributions $w_iP_i$: order by $w_i\|P_i\|$, use $V_0^w=\sum_iw_iP_i$, and the Gram becomes $G^w_{ij}=w_iw_j(h_i\cdot h_j)(\phi_i\cdot\phi_j+1)$.
Proposition~\ref{prop:bound} then applies to the ideal weighted raw-gradient criterion; the practical implementation has the same epsilon and rounding qualifications. Each retained loss is weighted once, and $w\equiv1$ recovers the unweighted mask.
The unweighted rule carries no such guarantee:

\begin{example}[Unweighted selection violates the budget; constructed]\label{ex:weighted}
In one dimension, let candidate $a=+0.3$ carry weight $2$, candidate $c=-0.3$ weight $1$, and the other rows contribute $+3.0$ with weight $1$; $b=0.05$.
Unweighted, $\{a,c\}$ sums to $0$, so both are dropped.
The applied gradient is $2(0.3)-0.3+3.0=3.3$ and the dropped part is $0.3$, a relative change of $0.3/3.3\approx0.091>0.05$.
Weighted, the order is $c$ ($0.3$), $a$ ($0.6$); both prefixes have relative distortion $0.3/3.3\approx 0.091$, so $k^*=0$.
\end{example}

\subsection{Re-weighting the dropped candidates}\label{sec:reweight}
Deterministic dropping biases the update: the skipped part $S_{k^*}$ is small at the last layer, but it is removed at every step.
We evaluate four corrections applied to the dropped set $R=\{\pi_1,\dots,\pi_{k^*}\}$; all use only forward quantities.
\begin{itemize}[leftmargin=*,itemsep=1pt,topsep=2pt]
\item \textbf{Uniform keep (rw).} Keep each $i\in R$ with probability $q$ and weight $1/q$ ($q=0.5$). Conditional on $R$, the expected raw gradient equals the dense gradient for \emph{every} layer before clipping or optimization; the expected back-propagated fraction is $1-(1-q)k^*/B$. The realised last-layer change is no longer bounded by $b$.
\item \textbf{Norm-proportional keep (rwp).} Keep $i\in R$ with $p_i=\mathrm{clip}(\|P_i\|/\tau,p_{\min},1)$ and weight $1/p_i$, with $\tau$ chosen numerically per batch to target mean probability $q$ over $R$ ($p_{\min}=0.05$, so weights are at most $20$). The mean is non-increasing in $\tau$, so $\tau$ is found by a device-only grid bisection in $\log\tau$ (4 rounds of 64 points); the target mean need not be attained exactly. Inverse-probability weighting remains unbiased for the raw gradient.
\item \textbf{Single rescale (det\_scale).} When both norms are nonzero, drop $R$ and multiply the kept rows' losses by $c=\|V_0\|/\|V_0-S_{k^*}\|$, restoring the last-layer gradient norm. For ideal selection with $V_0\ne0$ and $b<1$, $c\in[1/(1+b),\,1/(1-b)]$ by the triangle inequality.
\item \textbf{Last-layer fix (det\_lastfix).} Drop $R$ and add $S_{k^*}$ (computable as $(H\odot m_R)^\top\tilde\Phi$) to the final layer's gradient after the backward, so, in exact arithmetic, the final layer receives the dense gradient and deeper layers the selected one.
\end{itemize}

\section{Implementation}\label{sec:impl}

\paragraph{Numerical criterion.}
Algorithm~\ref{alg:lacr} uses $\epsilon=10^{-12}$ and fp32 Gram arithmetic, clamping negative squared norms from roundoff to zero. Ignoring rounding, its accepted prefix obeys $\|S_k\|\le b(\|V_0\|+\epsilon)$, rather than the ideal bound $b\|V_0\|$. For $V_0\ne0$, the relative bound is $b(1+\epsilon/\|V_0\|)$; at $V_0=0$, the absolute allowance is $b\epsilon$. Finite-precision evaluation can add error, so this is not a numerically certified exact-$b$ guarantee.

\paragraph{Factorised Gram.}
Since $P_a\cdot P_b=(h_a\cdot h_b)(\tilde\phi_a\cdot\tilde\phi_b)$, the Gram of the sorted candidates is $G=(h_\pi h_\pi^\top)\odot(\phi_\pi\phi_\pi^\top+1)$, and $\|S_k\|^2=\sum_{a,b\le k}G_{ab}$ is the $k$-th diagonal entry of the 2-D cumulative sum of $G$.
All $B+1$ prefix norms are obtained at once.
Row norms and the denominator also factorise, $\|P_i\|=\|h_i\|\sqrt{\|\phi_i\|^2+1}$ and $\|V_0\|^2=\|h^\top\phi\|_F^2+\|\sum_i h_i\|^2$, so $P$ need never be formed.
At $B=128$, $C=100$, $F=512$, the materialised $P$ is about 25\,MiB and its Gram about 1.7\,GFLOP, whereas the factorised Gram needs two small matrix products ($B^2C$ and $B^2F$ multiply--adds).
In the runs reported here the selector factorises the Gram (the dominant cost) and forms the row norms and denominator from $P$; a fully factorised variant gives the same mask in exact arithmetic.
In fp32 the two forms round differently and could flip a mask when some $D_k$ lies within rounding of $b$; the masks were bitwise identical to the reference on 1000 of 1000 stored batches (at $b=0.09$ and $0.25$) and over 300 training steps.

\paragraph{No host synchronisation in the selector.}
Candidate masking, the sort (non-candidates get key $+\infty$), the cumulative sums, the comparison with $b$ and $k^*=\max_k k\,[k\le|\mathcal{C}|][D_k\le b]$ are tensor operations; the drop mask stays on the device.
Logged statistics are accumulated on the device and read once per epoch.
The only remaining host read per step is the kept-row count that sizes the backward.
On an RTX 5090 at a kept fraction of $0.69$ the selector costs $0.12$\,ms per step, against $0.71$\,ms for the unfactorised Gram (of which $0.65$\,ms is the $B\times B$ GEMM); the factorised selector shortened the whole step by $5.4\%$ \ci{5.2}{5.6}.

\paragraph{Execution requirement.}
Selecting rows does not by itself save time.
If the rule is applied as a masked loss (dropped rows get zero loss weight), the backward still runs over all $B$ rows.
We therefore use a one-forward selective-backward execution that reuses the scoring forward's activations and back-propagates only the kept rows, normalised by $B$; it targets the gradient of the selected objective in exact arithmetic.
We checked the update against the masked-loss reference: the relative $L_2$ difference was $1.5\times10^{-6}$ in fp32 (TF32 off) for the weighted composition.
Under a cost model with forward $=1$ and backward $=2$ units, the ideal time ratio at kept fraction $k$ is $(1+2k)/3$.
Table~\ref{tab:exec} reports measured costs.
On CIFAR-100 ResNet-18, the masked-loss execution of the same rule took $1.072\times$ \ci{1.069}{1.075} the time of the activation-reusing execution at 200 epochs ($1.054\times$ at 120 epochs), with test accuracy differences of $-0.17$\pp{} \ci{-1.12}{+0.78} and $+0.41$\pp{} \ci{-0.54}{+1.36}, which does not establish numerical trajectory equivalence.
In the weighted InfoBatch composition the masked-loss version took $1.117\times$ \ci{1.116}{1.117} the time of the reusing execution and $1.016\times$ \ci{1.015}{1.018} the time of InfoBatch alone, i.e.\ no saving over InfoBatch despite back-propagating fewer rows.
The remaining overhead over $(1+2k)/3$ is dominated by gathering the kept rows' cached activations: $0.545$\,ms per step against a $0.542$\,ms memory-bandwidth roofline for the same bytes, suggesting limited benefit from scheduling changes alone for that measured gather.

\begin{table}[t]
\centering\small
\caption{Execution cost, CIFAR-100 ResNet-18-GN, $B=128$, $b=0.09$. Top: per-step stage times (ms) on an RTX 5090 from an epoch-60 training state (kept fraction $\approx0.69$). Bottom: whole-step and whole-run time ratios to dense on the same GPU; overhead $=$ ratio $-(1+2\bar k)/3$.}
\label{tab:exec}
\begin{tabular}{lccccc}
\toprule
stage (ms/step) & forward & selector & gather & backward & optimizer\\
\midrule
dense & 2.77 & -- & -- & 6.97 & 0.22\\
CAPSkip, reusing execution & 2.70 & 0.12 & 0.60 & 4.66 & 0.22\\
\bottomrule
\end{tabular}

\vspace{4pt}
\begin{tabular}{lccc}
\toprule
measurement & ratio to dense & $\bar k$ & overhead\\
\midrule
per-step timer, RTX 5090 (300 steps $\times$ 6 rounds) & 0.835 & 0.689 & $+0.042$\\
200-epoch run, RTX 5090 ($n=3$) & 0.888 \ci{0.887}{0.889} & 0.741 & $+0.061$\\
200-epoch run, RTX 4090 ($n=3$) & 0.917 \ci{0.910}{0.924} & 0.742 & --\\
masked loss / reusing execution, RTX 4090, 200 ep ($n=3$) & 1.072 \ci{1.069}{1.075} & -- & --\\
\bottomrule
\end{tabular}
\end{table}

\section{Experiments}\label{sec:exp}

\subsection{Protocol}
\paragraph{Statistics.}
Runs are paired by seed.
Accuracy differences are reported as mean and 95\% $t$-interval over seed pairs; time ratios and time-to-accuracy (TTA) ratios as the geometric mean of per-seed ratios with a 95\% $t$-interval in log space.
A difference is called better/worse (faster/slower) only if its interval excludes $0$ ($1$); otherwise the difference is unresolved; this is not a claim of equivalence.
Seed counts are small ($n=3$ or $5$), so power is low.
Times are training-loop wall-clock and are compared only within one machine.
Primary comparisons and decision rules were written down before the runs; tables mark secondary or exploratory rows.
Learning-rate and schedule choices were never revised after looking at test accuracy.

\paragraph{Fine-tuning.}
DeiT-S/16~\citep{touvron2021deit} pre-trained on ImageNet-1k, head replaced by a zero-initialised 100-way layer, fine-tuned on CIFAR-100~\citep{krizhevsky2009cifar} (50k train, 10k test) following the ViT transfer protocol~\citep{dosovitskiy2021vit} at $224\times224$: SGD with momentum $0.9$, no weight decay, base learning rate $0.03$ scaled by $B/512$, $B=128$, global-norm clipping at $1.0$, linear warm-up over the first 5\% of the run then cosine decay to $0$, no Mixup, no label smoothing (one-hot targets, so $t_i$ in \eqref{eq:P} is one-hot), drop path $0$, bf16 autocast, 103 epochs, seeds 17, 29, 43, one RTX 5090.
CAPSkip uses $b=0.09$ without tuning.
The approximately time-matched dense baseline uses $\mathrm{round}(103\,\hat\rho)=59$ epochs with the schedule compressed accordingly, where $\hat\rho=0.5763$ is the CAPSkip/dense training-time ratio of seed 17, fixed before the other seeds were run. Mean training times were 888.2\,s for CAPSkip103 and 883.1\,s for dense59, a 0.6\% mismatch.

\paragraph{From scratch.}
ResNet-18~\citep{he2016resnet} with GroupNorm~\citep{wu2018groupnorm} on CIFAR-100 with a 45k/5k train/validation split and the 10k test set; $B=128$, label smoothing $0.1$~\citep{szegedy2016inception}, one-cycle schedule~\citep{smith2019superconvergence} compressed to each epoch budget, weight decay $10^{-3}$, peak learning rate $0.1$ for dense and CAPSkip, $b=0.09$, final test accuracy.
For TTA, each arm is trained at several epoch budgets; per seed, the (time, final test accuracy) points form a running-max frontier, the time to reach a fixed target is linearly interpolated, and the per-seed ratio is CAPSkip/dense.

\subsection{Main result: DeiT-S fine-tuning on CIFAR-100}

\begin{table}[t]
\centering\small
\caption{DeiT-S fine-tuned on CIFAR-100 (test top-1, \%). Approximate time matching: CAPSkip at 103 epochs vs dense at 59 epochs (primary). Rows back-propagated are a percentage of forwarded rows. Means in the last row are computed from the per-seed values.}
\label{tab:deit}
\resizebox{\linewidth}{!}{%
\begin{tabular}{ccccccc}
\toprule
seed & dense 103 ep & dense 59 ep & CAPSkip 103 ep & CAPSkip $-$ dense 59 & CAPSkip $-$ dense 103 & rows bwd (\%)\\
\midrule
17 & 87.07 & 86.94 & 87.24 & $+0.30$ & $+0.17$ & 34.9\\
29 & 87.29 & 87.07 & 87.34 & $+0.27$ & $+0.05$ & 35.1\\
43 & 87.39 & 86.99 & 87.16 & $+0.17$ & $-0.23$ & 34.8\\
\midrule
mean [95\% CI] & 87.25 & 87.00 & 87.25 & $+0.25$ \ci{+0.08}{+0.42} & $-0.00$ \ci{-0.51}{+0.51} & \\
train time (s) & 1541 & 883 & 887--890 & & & \\
\bottomrule
\end{tabular}}
\end{table}

Table~\ref{tab:deit} gives the primary comparison.
At approximately matched training time CAPSkip is better than dense by $+0.25$\pp{} \ci{+0.08}{+0.42} ($n=3$).
At equal epochs, CAPSkip's accuracy difference from dense is unresolved ($-0.00$\pp{} \ci{-0.51}{+0.51}). It takes $0.576\times$ \ci{0.574}{0.579} the training time, back-propagating about 35\% of the forwarded rows.
The positive result compares more CAPSkip epochs with fewer dense epochs at approximately matched time; it does not establish a same-epoch accuracy gain.
Exploratory: the CAPSkip test curves (evaluated every 4 epochs) first reach the dense 103-epoch mean (87.25\%) at epochs 60, 72 and 64 (481, 592 and 517\,s); these intermediate points are not at the end of the learning-rate schedule and are not a decision.
Peak allocated memory was 5078\,MiB for CAPSkip against 4716\,MiB for dense, the cost of caching activations for reuse.

\begin{table}[t]
\centering\small
\caption{DeiT-S controls ($n=3$, 103 epochs each). Dense margin is accuracy minus the dense running-maximum envelope, linearly interpolated at each arm's own runtime using 30/59/80/103-epoch dense runs. The last column compares these margins, not accuracies at one common time. CAPSkip's interpolated margin is $+0.24$\pp{}, distinct from the direct dense59 contrast of $+0.25$\pp{}. Rows bwd is a percentage of forwarded rows.}
\label{tab:deit-ctrl}
\resizebox{\linewidth}{!}{%
\begin{tabular}{lcccc}
\toprule
arm & rows bwd & dense margin (pp) & time / top-1 vs CAPSkip 103 ep & margin contrast (pp)\\
\midrule
CAPSkip & $\approx$35\% & $+0.24$ \ci{+0.07}{+0.42} & -- & --\\
Selective-Backprop, same execution & 33.9\% & $-0.39$ \ci{-0.55}{-0.22} & 0.919 / $-0.71$ \ci{-1.08}{-0.34} & $+0.63$ \ci{+0.29}{+0.97}\\
InfoBatch ($r=0.5$) & 100\% & $-0.09$ \ci{-0.22}{+0.05} & 1.077 / $-0.29$ \ci{-0.44}{-0.14} & $+0.33$ \ci{+0.29}{+0.37}\\
random + rescale & 100\% & $+0.00$ \ci{-0.23}{+0.23} & 1.077 / $-0.20$ \ci{-0.65}{+0.25} & $+0.24$ \ci{-0.13}{+0.61}\\
InfoBatch $\to$ weighted CAPSkip & 37.6\% & $+0.29$ \ci{+0.02}{+0.56} & 0.647 / $-0.31$ \ci{-0.97}{+0.36} & $-0.05$ \ci{-0.49}{+0.40}\\
\bottomrule
\end{tabular}}
\end{table}

Table~\ref{tab:deit-ctrl} adds controls with the same recipe and seeds after the primary fine-tuning result.
For seed $s$, define $m_{a,s}=A_{a,s}-D_s(t_{a,s})$, where $D_s$ is the dense envelope described in the caption. The last column is the paired mean of $m_{\mathrm{CAPSkip},s}-m_{a,s}$.
Selective-Backprop~\citep{jiang2019selectivebackprop} ($\beta=2$) uses the same activation-reusing kernels but pools selected rows before stepping; CAPSkip steps after every forwarded batch with the original $B$ denominator. This compares policies together with their update protocols, rather than isolating the mask alone. SB back-propagates 33.9\% of forwarded rows and has a dense-relative margin of $-0.39$\pp{}.
InfoBatch uses ratio 0.5. Its paired random control draws the same number of candidates uniformly from the full training set, independently skips each with probability 0.5, doubles surviving candidate weights, and leaves other weights at one. It does not match CAPSkip's retained-row count. Outer pruning operates in epochs 2--90 and stops in epochs 91--103; the hybrid's inner CAPSkip continues throughout. SB uses one dense priming epoch; CAPSkip has no dense prime.
The CAPSkip--SB and CAPSkip--InfoBatch margin contrasts are $+0.63$\pp{} \ci{+0.29}{+0.97} and $+0.33$\pp{} \ci{+0.29}{+0.37}. These support the margin comparison, not the literal planned claim of direct superiority at one common runtime.
The random-control contrast is $+0.24$\pp{} \ci{-0.13}{+0.61}, so no difference is resolved. This leaves the benefit of the specific row-selection rule unresolved, although the control does not match CAPSkip's row counts or update protocol.
InfoBatch followed by weighted CAPSkip has a dense-relative margin of $+0.29$\pp{} \ci{+0.02}{+0.56}, takes $0.647\times$ CAPSkip's time, and has an unresolved margin difference from CAPSkip alone. Three seeds on one task do not establish generality. No same-rule re-forward CAPSkip control or equal-time accuracy comparison against compiled dense was run in this case.

\subsection{CIFAR-100 from scratch: no resolved time-to-accuracy difference}

\begin{table}[t]
\centering\small
\caption{ResNet-18-GN trained from scratch on CIFAR-100, CAPSkip ($b=0.09$) vs dense, RTX 4090, seeds 2--6 ($n=5$). Left: equal epochs. Right: test time-to-accuracy over the 100--200-epoch frontier.}
\label{tab:scratch}
\begin{tabular}{ccc}
\toprule
epochs & time ratio & test diff (pp)\\
\midrule
100 & 0.926 \ci{0.900}{0.954} & $-0.15$ \ci{-0.72}{+0.43}\\
120 & 0.928 \ci{0.920}{0.936} & $-0.38$ \ci{-0.92}{+0.16}\\
140 & 0.927 \ci{0.925}{0.929} & $-0.00$ \ci{-0.30}{+0.30}\\
160 & 0.920 \ci{0.913}{0.926} & $-0.17$ \ci{-0.69}{+0.35}\\
200 & 0.918 \ci{0.915}{0.921} & $-0.06$ \ci{-0.24}{+0.13}\\
\bottomrule
\end{tabular}\quad
\begin{tabular}{lcc}
\toprule
test target & TTA ratio & verdict\\
\midrule
74.978 (primary) & 1.053 \ci{0.921}{1.203} & unresolved\\
74.478 (expl.) & 0.992 \ci{0.873}{1.127} & unresolved\\
75.2 (expl., $n=4$) & 0.987 \ci{0.802}{1.215} & unresolved\\
\bottomrule
\end{tabular}
\end{table}

Table~\ref{tab:scratch} shows that per epoch, CAPSkip takes $0.918$--$0.928\times$ the time with every test-accuracy interval containing $0$, and that at the pre-registered test target the TTA ratio is $1.053$ \ci{0.921}{1.203}, resolving no time-to-accuracy difference.
The target was fixed earlier from a different machine's dense runs; dense on this machine averages 75.35\% at 200 epochs, above the target, and the target was not moved.
The time saving per epoch is small for two reasons visible in Table~\ref{tab:exec}: CAPSkip keeps about 74\% of the rows at 200 epochs (versus 35\% in fine-tuning), and the fixed per-step overheads (gather, selector) are a larger fraction of a ResNet-18 step.

An earlier five-seed experiment on a different GPU with a partly different configuration also resolved no test-target difference; its validation-target result is discussed in Appendix~\ref{app:details}.

\subsection{Composition with InfoBatch}\label{sec:exp-ib}

\paragraph{Setup.}
Same from-scratch recipe, seeds 2--6, InfoBatch with $r=0.5$ and a full-data tail in the last 12.5\% of epochs.
In this recipe InfoBatch at peak learning rate $0.1$ without clipping collapsed transiently in every run we made (validation accuracy falling to as low as 1\% mid-training), which we traced to tiny, up-weighted final batches at high learning rate; tuning InfoBatch alone on seed-1 validation selected peak learning rate $0.05$ with gradient clipping at norm $5$, which removed the collapses.
All InfoBatch-family arms use this setting.
The primary comparison (fixed in advance) is each composite arm against InfoBatch alone on the same machine at 200 epochs, and ``improves InfoBatch'' requires a time-ratio upper bound $<1$ \emph{and} a test-difference lower bound $>-0.3$\pp.
The first experiment ran one machine per seed (four rented RTX 5090 instances and one local RTX 5090 workstation); the replication ran all five seeds on the local workstation (its seed-6 runs are shared with the first experiment, so the two sets are not fully independent).

\begin{table}[t]
\centering\small
\caption{InfoBatch composition, CIFAR-100 ResNet-18-GN, 200 epochs, $n=5$ unless noted. Time ratio / test difference vs InfoBatch on the same machine. Removal $=1-$ rows back-propagated $/45{,}000$ per epoch, averaged over epochs, taken from the first experiment's runs.}
\label{tab:ib}
\resizebox{\linewidth}{!}{%
\begin{tabular}{llccc}
\toprule
& arm & vs InfoBatch: time / test (pp) & bar passed & removal\\
\midrule
\multirow{4}{*}{\shortstack[l]{one machine\\per seed}}
& unweighted CAPSkip inside IB & 0.918 \ci{0.908}{0.929} / $-0.52$ \ci{-0.86}{-0.19} & no & 0.508\\
& \textbf{weighted CAPSkip inside IB} & 0.924 \ci{0.915}{0.934} / $-0.04$ \ci{-0.26}{+0.19} & \textbf{yes} & 0.503\\
& weighted, up-weighted rows excluded & 1.048 \ci{1.037}{1.060} / $+0.06$ \ci{-0.37}{+0.49} & no (time) & 0.401\\
& IB first half, CAPSkip second half & 1.084 \ci{1.076}{1.092} / $+0.29$ \ci{-0.12}{+0.69} & no (time) & 0.367\\
\midrule
\multirow{4}{*}{\shortstack[l]{single\\machine}}
& \textbf{weighted CAPSkip inside IB} & 0.910 \ci{0.908}{0.912} / $-0.14$ \ci{-0.45}{+0.17} & \textbf{no (acc.)} & --\\
& same, masked-loss execution & 1.016 \ci{1.015}{1.018} / $-0.30$ \ci{-1.02}{+0.43} & no & 0.502\\
& IB throughout, CAPSkip added in 2nd half & 0.918 \ci{0.917}{0.920} / $-0.31$ \ci{-0.73}{+0.11} & no (acc.) & 0.441\\
& IB first half, $b{=}1$ + rw second half ($n=4$) & 1.068 \ci{1.064}{1.072} / $-0.31$ \ci{-0.60}{-0.02} & no & 0.372\\
\midrule
& InfoBatch alone & -- & -- & 0.312\\
\bottomrule
\end{tabular}}
\end{table}

\paragraph{Results.}
Table~\ref{tab:ib} reports both experiments.
Weighting matters: with the same removal rate ($\approx0.50$), the unweighted inner rule loses $0.52$\pp{} \ci{-0.86}{-0.19} against InfoBatch while the weighted rule is within $-0.04$\pp{} \ci{-0.26}{+0.19}, as Example~\ref{ex:weighted} predicts.
In the first experiment the weighted composite passes the pre-registered bar ($0.924\times$ the time).
In the single-machine replication it is again faster ($0.910\times$) but its accuracy lower bound, $-0.45$\pp, misses the $-0.3$\pp{} bar.
The speed-up is therefore reproducible, but the pre-specified accuracy tolerance is met in only one of two experiments, and we do not claim that CAPSkip improves InfoBatch.
Without activation reuse the composite saves nothing ($1.016\times$ InfoBatch).
Time-switched variants did not help.
Deviations from the plan are listed in Appendix~\ref{app:details}.

\paragraph{Learning-rate confound.}
Because the InfoBatch family runs at learning rate $0.05$ with clipping while dense used $0.1$, we also ran dense at the InfoBatch setting.
It is $+0.59$\pp{} \ci{+0.22}{+0.96} better than dense at $0.1$ at 200 epochs ($+0.47$ \ci{+0.23}{+0.72} at 120 epochs; $n=5$); on the one seed run on the same machine its time ratio was $1.007$, i.e.\ no time difference.
Any apparent accuracy edge of the InfoBatch family over dense at learning rate $0.1$ (InfoBatch alone: $+0.43$\pp{} \ci{+0.00}{+0.85}) is mostly this learning-rate effect.
Against matched-HP dense on the same machine at 200 epochs, InfoBatch takes $0.696\times$ \ci{0.695}{0.697} the time at $-0.05$\pp{} \ci{-0.64}{+0.54}, and weighted CAPSkip inside InfoBatch $0.633\times$ \ci{0.632}{0.634} at $-0.19$\pp{} \ci{-0.70}{+0.32}; both accuracy differences are unresolved.
Accuracy-only comparisons against matched-HP dense over all five seeds are in Appendix~\ref{app:details}.

\subsection{Re-weighting variants and budget saturation}

\begin{table}[t]
\centering\small
\caption{Left: kept fraction $\bar k$ (rows back-propagated / forwarded, mean over a 30-epoch run, seed 1, RTX 4090) vs budget $b$. Right: re-weighting of the dropped candidates at $b=1.0$ (warm-up of 3 epochs, CAPSkip off in the last 12.5\%), mean final \emph{validation} accuracy at 200 epochs over seeds 2--4, and training time on rented RTX 5090 machines.}
\label{tab:rw}
\begin{tabular}{cc}
\toprule
$b$ & $\bar k$\\
\midrule
0.09 & 0.848\\
0.15 & 0.810\\
0.25 & 0.763\\
0.4 & 0.695\\
0.6 & 0.631\\
1.0 & 0.610\\
\bottomrule
\end{tabular}\qquad
\begin{tabular}{lcc}
\toprule
weighting & val acc (\%) & train time (s)\\
\midrule
det (drop only) & 73.893 & 589.0\\
rw ($q=0.5$) & \textbf{75.107} & 685.8\\
rwp ($q=0.5$, $p_{\min}=0.05$) & 74.600 & 740.4\\
det\_scale & 73.520 & 593.3\\
det\_lastfix & 73.567 & 598.3\\
\bottomrule
\end{tabular}
\end{table}

\paragraph{Saturation.}
Table~\ref{tab:rw} (left) shows that $\bar k$ flattens above $b\approx0.6$: only correctly classified rows are candidates, so a larger budget cannot remove more than the correct rows, and no grid value reached $\bar k\le0.50$.

\paragraph{Re-weighting.}
At $b=1.0$, deterministic dropping (without warm-up or annealing) costs accuracy: against $b=0.09$ it is $0.794\times$ \ci{0.767}{0.823} the time but $-3.75$\pp{} \ci{-4.30}{-3.20} in test accuracy at 200 epochs.
Among the corrections (Table~\ref{tab:rw}, right; selected on validation), uniform keep-and-reweight (rw) is best; on test it is $+1.05$\pp{} \ci{-0.09}{+2.19} above drop-only at 200 epochs ($+0.86$ \ci{+0.08}{+1.64} at 120) at $1.164\times$ \ci{1.134}{1.196} its time, because it back-propagates more rows ($\bar k$ $0.759$ vs.\ $0.571$).
The two deterministic corrections (det\_scale, det\_lastfix) do not show a resolved accuracy improvement; these data do not establish that either correction recovers the loss at large $b$.
In the end rw at $b=1$ lands where deterministic $b=0.09$ already is: $0.899\times$ \ci{0.885}{0.913} dense time at $-0.21$\pp{} \ci{-0.81}{+0.38}, against $0.903\times$ \ci{0.893}{0.912} at $-0.17$\pp{} \ci{-0.58}{+0.24} on the same machines ($n=3$).
Warm-up, final annealing and a linear budget ramp changed accuracy only within seed noise.

\section{Related work}

\paragraph{Per-step selection.}
Online batch selection~\citep{loshchilov2015online} and Selective-Backprop~\citep{jiang2019selectivebackprop} back-propagate high-loss examples; importance sampling~\citep{zhao2015importance,katharopoulos2018notall} samples by (an upper bound on) the per-example gradient norm and re-weights to stay unbiased, and \citet{katharopoulos2018notall} also use the gradient with respect to the network output as a cheap proxy.
RHO-LOSS~\citep{mindermann2022rholoss} selects by reducible held-out loss.
VCAS~\citep{wang2024vcas} samples rows and tokens inside the backward with controlled variance.
These methods use different selection or sampling objectives. CAPSkip evaluates the skipped \emph{set} through its summed last-layer gradient, with an ideal deterministic bound rather than an unbiased-gradient claim.
Closest to our setting, \citet{balles2023negative} select, within each forwarded batch of Selective-Backprop, a weighted subset whose last-layer gradient matches the batch mean, and report that neither this nor loss-based selection consistently beats a random subset; we include a matched random control for the same reason (\S\ref{sec:exp}).
K-ABENA~\citep{bonbhel2026kabena} excludes low-loss samples from the backward and keeps an unbiased estimate by inverse-probability re-weighting, which is related to our stochastic re-weighting variants (\S\ref{sec:reweight}) rather than to the deterministic budget.
For LLM fine-tuning, LOOM~\citep{chen2026loom} selects, at every step, a non-negatively weighted subset of the batch that matches a leave-one-out gradient target in the optimizer's metric; it reports that allowing signed (cancelling) weights lowered gradient error but hurt accuracy, whereas CAPSkip uses cancellation only to decide which rows to \emph{omit}, with weight one on all kept rows.

\paragraph{Per-epoch pruning.}
InfoBatch~\citep{qin2024infobatch} skips low-loss examples with probability $r$ and rescales the rest, with a full-data tail; KAKURENBO~\citep{nguyen2023kakurenbo} hides examples adaptively and adjusts the learning rate; RS2~\citep{okanovic2024rs2} shows that repeated random subsets are a strong time-to-accuracy baseline.
CAPSkip is complementary: it saves backward work on rows that were forwarded anyway, and \S\ref{sec:weighted} gives the form that keeps its bound when combined with such re-weighting pruners.

\paragraph{Gradient matching and coresets.}
CRAIG~\citep{mirzasoleiman2020craig} and GRAD-MATCH~\citep{killamsetty2021gradmatch} choose weighted subsets whose gradient sum approximates the full gradient, typically with last-layer gradient proxies and periodic re-selection.
CAPSkip uses the same insight, that cancellation within the sum matters, in the reverse direction and at every step: it chooses what to \emph{omit} so that the omitted sum is small, using exact last-layer gradients from the current forward rather than stale proxies.
Last-layer gradient embeddings $h_i\otimes\tilde\phi_i$ also appear in active learning~\citep{ash2020badge}, and score-based static pruning uses per-example gradient or error norms~\citep{paul2021datadiet,coleman2020svp}.
LESSER~\citep{zhang2026lesser} uses output-layer gradients, which need only a forward pass, as features for ranking post-training data by alignment with validation gradients; CAPSkip uses the same layer's gradients online, within each batch, and through their sum rather than their individual alignment.
The factorisation $\|P_i\|=\|h_i\|\|\tilde\phi_i\|$ is the standard per-example gradient-norm trick for linear layers~\citep{goodfellow2015perexample}; our use of the factorised \emph{Gram} for all prefix norms at once is what makes the full scan cheap.

\section{Limitations}
\begin{itemize}[leftmargin=*,itemsep=1pt,topsep=2pt]
\item \textbf{Evidence.} The only approximately time-matched accuracy gain over dense training is one fine-tuning task with three seeds. From-scratch training resolves no time-to-accuracy difference, and the InfoBatch composition passes its bar in one of two experiments. We have not tested larger datasets, language models, or other optimisers.
\item \textbf{Scope of the guarantee.} The ideal exact-arithmetic bound covers the raw final-affine-layer gradient at each step; the implementation is epsilon-relaxed and subject to rounding. It gives no control of deeper-layer gradients or of the trajectory; the accuracy drop at $b=1$ (\S\ref{sec:exp}) shows that the last-layer budget does not by itself protect accuracy, and the deterministic last-layer corrections do not recover it.
\item \textbf{Maximality is relative to an order.} $k^*$ is the longest feasible prefix of the norm-ascending order, not the largest feasible subset.
\item \textbf{Execution and memory.} Our activation-reusing execution saves time relative to masked-loss execution in the measured comparisons. This does not establish that reuse is necessary or superior to every alternative execution. Activation reuse costs memory (5078 vs.\ 4716\,MiB peak in our fine-tuning runs) and a bandwidth-bound gather.
\item \textbf{Budget saturation.} Candidates are limited to correctly classified rows, so the budget saturates and cannot reach low kept fractions; stochastic re-weighting trades the deterministic bound for unbiasedness.
\item \textbf{Hyper-parameters.} $b$ was tuned on validation in the from-scratch setting, and CAPSkip had more tuned configurations there than dense. The InfoBatch family was tuned separately, which is why we report comparisons against dense at the matched learning rate.
\end{itemize}

\section{Conclusion}
Measuring the summed, signed gradient of the examples one intends to skip, at the layer where it is exact, gives a selection rule with a simple per-step guarantee and an inexpensive on-device implementation.
Its practical value depends on the workload and on an execution that turns skipped rows into skipped work: in our experiments it improves on approximately time-matched eager dense in one fine-tuning case, resolves no time-to-accuracy difference from scratch, and, in weighted form, composes with InfoBatch for a reproducible speed-up whose accuracy cost is not yet resolved.

\bibliographystyle{plainnat}
\bibliography{refs}

\appendix
\section{Additional details}\label{app:details}

\paragraph{Earlier from-scratch experiment (validation targets).} An earlier five-seed experiment on an RTX 5090 with \emph{validation} targets reported a TTA ratio of $0.695$ \ci{0.618}{0.780} at the primary validation target (74.70\%), but $0.976$ \ci{0.826}{1.153} (unresolved) at the corresponding test target 74.978\%.
We do not count the validation result as evidence: the configurations of both arms were tuned on the same validation split, CAPSkip had more tuned configurations than dense (19 vs.\ 7), and that frontier changed configuration between budgets below and above 65 epochs.
The held-out test comparisons in both experiments resolve no difference.

\paragraph{InfoBatch composition: further results.} Accuracy-only comparisons against matched-HP dense across all five seeds are: InfoBatch $-0.17$ \ci{-0.68}{+0.34}, unweighted composite $-0.69$ \ci{-1.13}{-0.25}, weighted composite $-0.20$ \ci{-0.88}{+0.48}, up-weighted rows excluded $-0.11$ \ci{-0.39}{+0.17}, IB-then-CAPSkip $+0.12$ \ci{-0.23}{+0.47}. (pp, 200 epochs, $n=5$; seeds 2--5 of matched-HP dense ran on a different machine from the arms, so these are accuracy only.) Two deviations: test accuracy of partial results was seen twice while the queues were running (no arm, seed or rule was changed afterwards), and a spending cap on the rented machines moved three arms to the replication.

\paragraph{Code-level definition.} The candidate set, the sort with key $+\infty$ for non-candidates, the prefix-norm computation through a 2-D cumulative sum, and $k^*=\max_k k\,[k\le|\mathcal{C}|][D_k\le b]$ follow Algorithm~\ref{alg:lacr}; the denominator adds $\epsilon=10^{-12}$. In fine-tuning, $\phi_i$ is the class token after the final LayerNorm (the head's input) and targets are one-hot; in from-scratch training targets are label-smoothed with $\varepsilon=0.1$, matching the trained loss.

\paragraph{Fine-tuning execution.} The kept rows' backward runs in chunks of 64, accumulating gradients for one step per scoring batch with denominator $B=128$ and no cross-forward pooling. Chunking targets the same mathematical objective but may change floating-point reduction order. Approximate time matching uses training-loop time; whole-job times were 945, 945 and 1008\,s for the three CAPSkip runs versus 912--913\,s for dense at 59 epochs.

\paragraph{Fine-tuning implementation gate.} The DeiT check initially failed a fixed relative-$L_2$ gradient threshold of $2\times10^{-5}$. After that failure, a reference-relative diagnostic used $10\max(f,10^{-7})$, where $f$ is the reference discrepancy for each tensor and update; the largest error-to-floor ratio was 3.42. This post-hoc diagnostic covered ten early updates retaining 127--128 of 128 rows, rather than the roughly 35\% retention of the full run. It does not replace the failed fixed gate or validate the whole trajectory. The separate fp32 weighted-composition check reported in Section~\ref{sec:impl} has a different model and scope.

\paragraph{InfoBatch collapse diagnosis.} At peak learning rate 0.1 without clipping, InfoBatch-family runs had 60 of 222 epochs whose final batch had 1--8 rows end with validation accuracy halved, versus 2 of 4219 other epochs; dense and CAPSkip (whose final batch always has 72 rows) had 0 of 14023. With learning rate 0.05 and clipping at 5 there were 0 halvings in 2880 epochs.

\paragraph{Weighted-selector checks.} With all weights equal to 1, the weighted selector reproduced the unweighted masks on 200 of 200 batches, and its $k^*$ matched a float64 recomputation on 200 of 200 batches; with the exclusion variant, no row with weight $>1$ was ever dropped.

\end{document}